\documentclass[letterpaper, 10 pt, conference]{ieeeconf}  

\IEEEoverridecommandlockouts                              

\title{\LARGE \bf
Sector Patch Embedding: An Embedding Module Conforming to The Distortion Pattern of Fisheye Image
}

\author{Dianyi Yang$^{1,2}$, Jiadong Tang$^{1,2}$, Yu Gao$^{1,2}$, Yi Yang$^{*,1,2}$, Mengyin Fu$^{1,2}$
\thanks{$^{1}$School of Automation, Beijing Institute of Technology, Beijing, China}%
\thanks{$^{2}$State Key Laboratory of Intelligent Control and Decision of Complex
System, Beijing Institute of Technology, Beijing, China}%
\thanks{*Corresponding author: Y. Yang Email: yang\_yi@bit.edu.cn}%
}

\usepackage{graphicx} 
\usepackage{float} 
\usepackage{subfigure} 
\usepackage{cite}
\usepackage{color}
\usepackage{url}
\usepackage{booktabs}
\usepackage{hyperref}
\usepackage{footnote}
\usepackage{threeparttable}

\begin{document}
\maketitle
\thispagestyle{empty}
\pagestyle{empty}

\begin{abstract}
Fisheye cameras suffer from image distortion while having a large field of view(LFOV). And this fact leads to poor performance on some fisheye vision tasks. One of the solutions is to optimize the current vision algorithm for fisheye images. However, most of the CNN-based methods and the Transformer-based methods lack the capability of leveraging distortion information efficiently. In this work, we propose a novel patch embedding method called Sector Patch Embedding(SPE), conforming to the distortion pattern of the fisheye image. Furthermore, we put forward a synthetic fisheye dataset based on the ImageNet-1K and explore the performance of several Transformer models on the dataset. The classification top-1 accuracy of ViT and PVT is improved by 0.75$\%$ and 2.8$\%$ with SPE respectively. The experiments show that the proposed sector patch embedding method can better perceive distortion and extract features on the fisheye images. Our method can be easily adopted to other Transformer-based models. Source code is at \textcolor[rgb]{0,0,1}{\url{https://github.com/IN2-ViAUn/Sector-Patch-Embedding}}.
\end{abstract}

\section{INTRODUCTION}

Fisheye cameras are widely used in autonomous driving, video surveillance, and virtual reality due to the large field of view. Compared with standard images based on the pinhole model, fisheye images are characterized by radial distortion. There are two main solutions to deal with the radial distortion of fisheye images, one is to calibrate the fisheye images to standard images and then performing vison tasks, which requires numerous computational resources. The second is to perform vision tasks such as classification, detection and segmentation directly on fisheye images\cite{index-1,index-2,index-3}. However, most existing computer vision algorithms are based on standard images. Adopting the algorithms to fisheye images directly may not acquire desirable results\cite{index-4,index-5}. Therefore, it is necessary to design optimization methods that adapt to the distortion of fisheye images.

Current vision tasks for fisheye images, such as object detection and classification, are mainly based on Convolutional Neural Networks(CNN) and vision transformers. CNN is effective both in collecting local features of images and learning the relationship of different scales. However, the spatial information mapped in fisheye images is different from standard images, which introduces external radial distortion and brings new challenges to conventional CNNs. Moreover, the different distances of pixels from the center in fisheye images lead to varying degrees of distortion, thus undermining the spatial invariance capability of CNN\cite{index-6}. In other words, CNNs with the fixed shape of kernels are not suitable for learning the pixel distribution patterns of distorted objects on fisheye images theoretically. Transformer constructs a series of tokens by segmenting each image into patches with positional embeddings, and then extracts features using encoder-based blocks\cite{index-7}. The unique structure of the performr enables it to achieve long-range dependency and have stronger global information integration capabilities compared to CNN\cite{index-8}. However, most Transformer models cannot adequately account for the distortion and symmetry in fisheye images. Specifically, the distortion is inconsistent for each patch, thus Transformer cannot perceive and utilize distortion distribution well\cite{index-9}. Furthermore, both CNN and Transformer operate the whole image. The invalid information on the boundary not only occupies computational resources but also has a particular impact on the model's performance.

To address the problems above, we propose a sampling method with a well-designed patch called Sector Patch Embedding (SPE). SPE is more suitable for the distortion pattern of fisheye images, and can be embedded into model that supports serialized inputs. The method takes advantage of the symmetry and improves the perception capability of the model as well. In addition, we design a polar coordinate encoding mode, which is compatible with SPE. Experimental results demonstrate that SPE can extract semantic features on fisheye images more effectively.

The main contributions of our work are the following:
\begin{itemize}

\item A sampling method with a well-designed patch named Sector Patch Embedding (SPE) is proposed. The application of SPE in Transformer-based models for fisheye image classification tasks obtains a better accuracy.

\item We construct a synthetic fisheye dataset from ImageNet-1K which has the same size as the original dataset.

\end{itemize}

The rest of the paper is organized as follows. Section II gives an overview of the fisheye dataset and optimization for fisheye distortion in vision tasks. Section III describes how we produced the synthetic fisheye dataset and explains our sector patch embedding method in detail. Section IV presents the experiment results. We conclude with a summary and outlook for future research in Section V.

\section{RELATED WORK}

\begin{figure*}
    \centering
    \includegraphics[width=\textwidth]{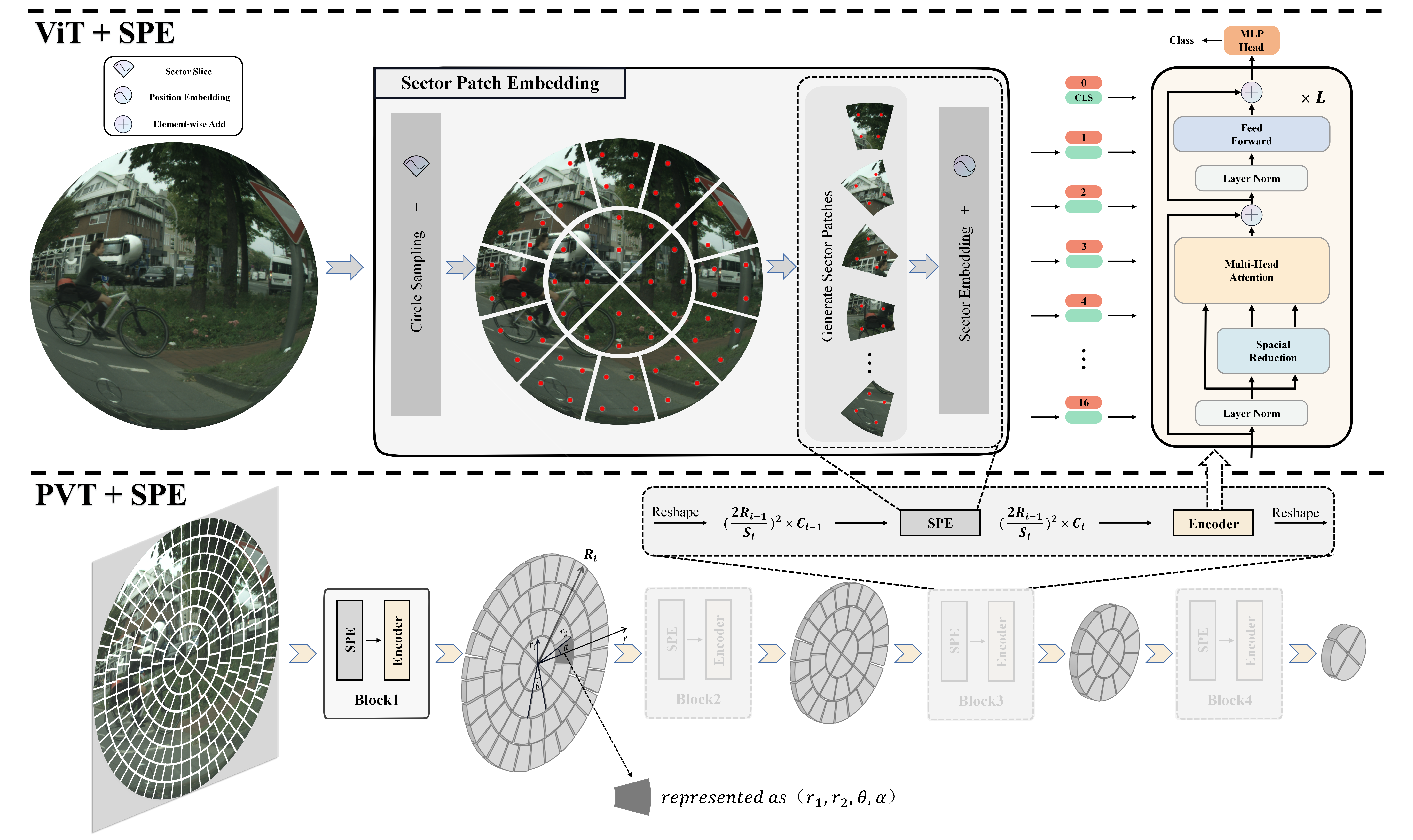} 
    \caption{\raggedright \textbf{Overview of Model Architecture.} This figure shows the \textbf{ViT} and \textbf{PVT} models with the introduction of the Sector Patch Embedding (\textbf{SPE}) module. We follow the original design of the Transformer encoder for both models. It is worth noting that the PVT model introduces Spatial Reduction Attention (\textbf{SRA}) to replace Multi-head Attention (\textbf{MHA}) in ViT, but when the reduction ratio in SRA is set to 1, it becomes MHA. So here we use Transformer layer under PVT to represent encoder module. The biggest feature of the improved model is the use of the SPE module to generate sector patches before the data is input to the Transformer encoder. For more model details, please refer to the \textit{Model architecture} section in Part III.}
    \label{fig1}
\end{figure*}

\subsection{Fisheye datasets}

Attributed to the labor-intensive and costly construction of fisheye datasets, they are far inferior to the standard image datasets in terms of quality and quantity. Specifically, as the first large-scale fisheye dataset for early use in autonomous driving, WoodScape \cite{index-10} realizes the omnidirectional perception of vehicles by four fisheye cameras and it accomplishes nine vision assignments including object detection and depth estimation. The dataset consists of over 10,000 images with semantic instance annotation and over 100,000 images with annotations for other tasks. Generally speaking, the advent of WoodScape significantly inspires research on fisheye images. Subsequently, the authors of WoodScape released SynWoodScape\cite{index-11}, which supports more than a dozen vision tasks. SynWoodScape addresses the weakness of WoodScape by simultaneously labeling four fisheye cameras. Thus it can obtain a bird-eye view perception through multi-camera algorithms. In the meanwhile, SynWoodScape provides pixel-level ground truth of optical flow and depth information for accurate and dense perception of near-field regions. KITTI-360\cite{index-12} complements the KITTI\cite{index-4} dataset with richer visual annotations and application scenarios. However, the downside is that KITTI-360 does not provide semantic annotations for the images captured by the two fisheye lenses. FisheyeCityScapes\cite{index-5} introduced a method of creating a synthetic fisheye dataset. Specifically,  FisheyeCityScapes simulate the fisheye images in any orientation at different focal lengths to achieve data augmentation. However, extensive noisy pixels are generated due to the interpolation algorithms, causing the images inconsistent with the distortion pattern of real fisheye images. OmniScape\cite{index-13} capture photorealistic environmental data in autonomous driving simulators and games via fisheye cameras. This approach can be extended to any virtual environment. Similar to OmniScape, THEODORE\cite{index-14} is established by simulating a fisheye camera in a virtual environment. Notably, THEODORE is a large-scale indoor dataset containing more than 100,000 high-resolution fisheye images. However, the camera’s perspectives of THEODORE are relatively homogeneous, all from an overhead view.

\begin{figure*}
    \centering
    \includegraphics[width=\textwidth]{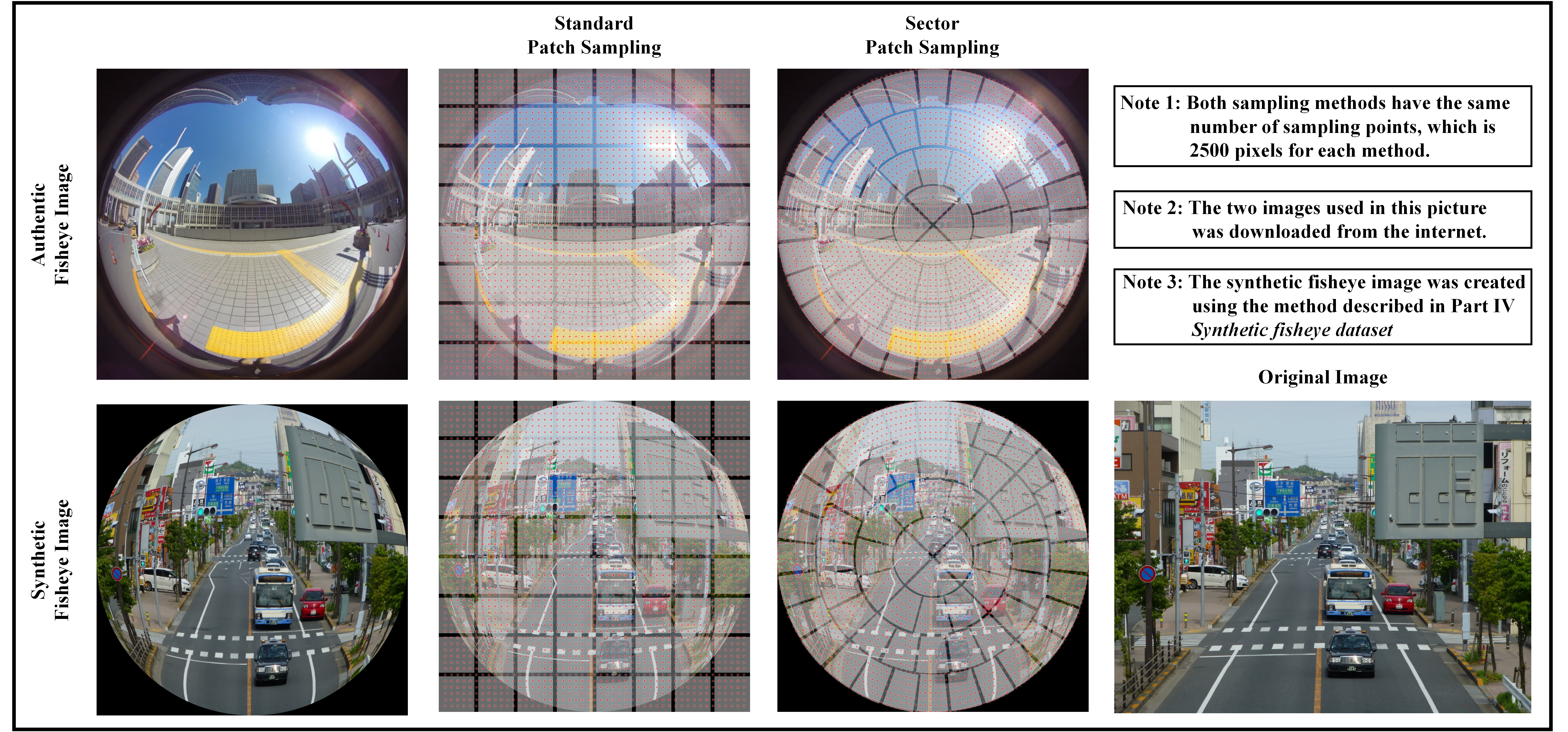}
    \caption{\textbf{Standard patch sampling and sector patch sampling on authentic (top) and synthetic fisheye image (bottom)}}
    \label{[patches]}
\end{figure*}

\subsection{Optimization for the distortion in fisheye images}

As fisheye cameras are widely applied in various fields, including autonomous driving and video surveillance, extensive research has been made which focuses on the optimization of fisheye images in vision tasks. FisheyeDet\cite{index-15} can extract image features adaptively without a priori information, such as camera parameters. Considering that the rectangular bounding box of object detection tasks in the standard image is not applicable to distorted objects in fisheye images, the authors designed an irregular quadrilateral bounding box regression strategy to achieve accurate and robust localization of distorted objects in fisheye images. The authors also propose a regular quadrilateral bounding box regression strategy to achieve accurate and robust localization of distorted objects in fisheye images. The method achieved 74.87$\%$ mAP on their own VOCFisheye dataset. SphereNet\cite{index-2} adjusted the sampling locations during convolution based on the principle of equidistant projection, transferring partial CNN operations, such as convolution and pooling, from typical 2D images to spherical surfaces. This enabled existing detection algorithms to be employed for spherical distorted objects. Saez et al.\cite{index-16} improved the convolution framework of the Efficient Residual Factorized Network(ERFNet)\cite{index-17}, enhancing the network's ability to handle distorted textures and making real-time accurate semantic segmentation available on synthetic fisheye images. It demonstrated promising results on real fisheye data. Goodarzi\cite{index-18} adapted SqueezeDet\cite{index-19} by adding a fisheye effect filter to the detection pipeline to accommodate the distortion. The fisheye enhancement technique proposed by the authors can be used in CNN-based fisheye image detection model. Song et al.\cite{index-20} improved the detection network YOLOX-tiny for the tree height detection task by adding an attention module for the global perception of image distortion. This improves detection efficiency while ensuring detection accuracy. FishFormer\cite{index-9} applies the Transformer structure to fisheye image correction, while the traditional correction method does not take the distortion of fisheye images into account. FishFormer cuts the fisheye image into annular patches to ensure the consistency of the distortion for each patch. The modification of the structure improves the perceptual capability of the model from both global and local perspectives. Zhang et al.\cite{index-21} employed a Spatial Transformer Layer to correct the feature maps obtained from Swin Transformer\cite{index-22}. Subsequently, feature maps are sent to PANet\cite{index-23} and the detection heads of R-CNN for detecting densely packed commodities in fisheye images. FisheyeYOLO\cite{index-24} explored representations for fisheye images relying on the radius distortion of fisheye images. They propose a polygonal object boundary box that adapts to the curvature of the object contours. This work increases the average loU relative accuracy by 40.3$\%$.

\section{METHOD}

\subsection{Model architecture}

The overall structure of the model is demonstrated in Fig.\ref{fig1} Based on the advantages of the Transformer mentioned in Part I, we choose Transformer-based models to extract the semantic features of the fisheye image. In order to adapt to the distortion pattern of fisheye images, we propose a new sampling method and the corresponding patch shape, which we call Sector Patch Embedding (SPE). To evaluate the proposed sampling method and patch shape, two representative Transformer-based models, ViT\cite{index-7} and PVT\cite{index-25}, are selected. ViT is a uniform-scale model with a multi-head attention mechanism, while PVT is a multi-scale model. Both ViT and PVT belong to the Pure-Transformer branch and have achieved excellent results in classification tasks. Other than the difference in training scale, the biggest difference between the two lies in the design of the attention module. We use SRA to represent the attention mechanism of PVT, and MHA to represent the attention mechanism of ViT. Given a query $Q$, a key $K$ and a value $V$ as input, the following formulas calculate the results of MHA and SRA respectively.

$$
{Attention}(\mathbf{q}, \mathbf{k}, \mathbf{v})={Softmax}\left(\frac{\mathbf{q k}^{\top}}{\sqrt{d_{head}}}\right) \mathbf{v} \eqno{(1)}
$$
$$ 
{MHA} ={Attention}\left(Q W^{Q}, K W^{K}, V W^{V}\right) \eqno{(2)}
$$
$$ 
{SRA}={Attention}\left(Q W^{Q}, \mathrm{SR}(K) W^{K}, \mathrm{SR}(V) W^{V}\right) \eqno{(3)}
$$

For the purpose of direct comparison, we only consider the calculation method under single-head attention. Equation(1) shows the attention calculation mechanism in the original Transformer\cite{index-26}. The difference between them is that SRA has an additional $SR(\cdot)$ parameter, which could be written as:
$$
SR(x) = Norm(Reshape(x, r)W^R) \eqno{(4)}
$$

where $x \in {R}^{(H \times W) \times C} $ represents input sequence, $r$ is the reduction ratio. After $Reshape(\cdot)$, the dimension of x becomes $\frac{HW}{r^2}\times(r^2C)$. Finally, through a linear projection with the weight matrix $W^R$, the dimension is reduced to $C$. The final output is reduced to ${R}^{\frac{HW}{r^2} \times C}$ from the original dimension.

To better integrate the SPE module into two models, we modify the existing models. For ViT, we sample the input fisheye image in a circular shape using bilinear interpolation, and generate sector patches. Each patch is reshaped into a d-dimensional vector, combined with positional encoding information, and then fed into the L-layer Transformer Encoders. Finally, the output is passed through an MLP for the final prediction.

Different from ViT, PVT generates new patches at different scales. Notably, we sample the original image in the first block only; subsequent blocks generate patches directly without sampling from the feature maps. To accommodate the multi-scale feature map generation, we use polar coordinates to store the sampling points and enable reshaping the circular image. In the first block, we generate sampling points and divide the image into patches, using $R_{i-1}$ to represent the maximum radius of the previous feature map and $s_i$ to represent the number of sampling points along the radius direction in each patch. In subsequent blocks, we reshape the input feature map, embed patches and positions using SPE, feed it to the Encoder, and reshape the output to obtain the feature map at the $R_i$ scale.

\subsection{The method of sampling}

\begin{figure}
    \centering
    \includegraphics[width=0.5\textwidth]{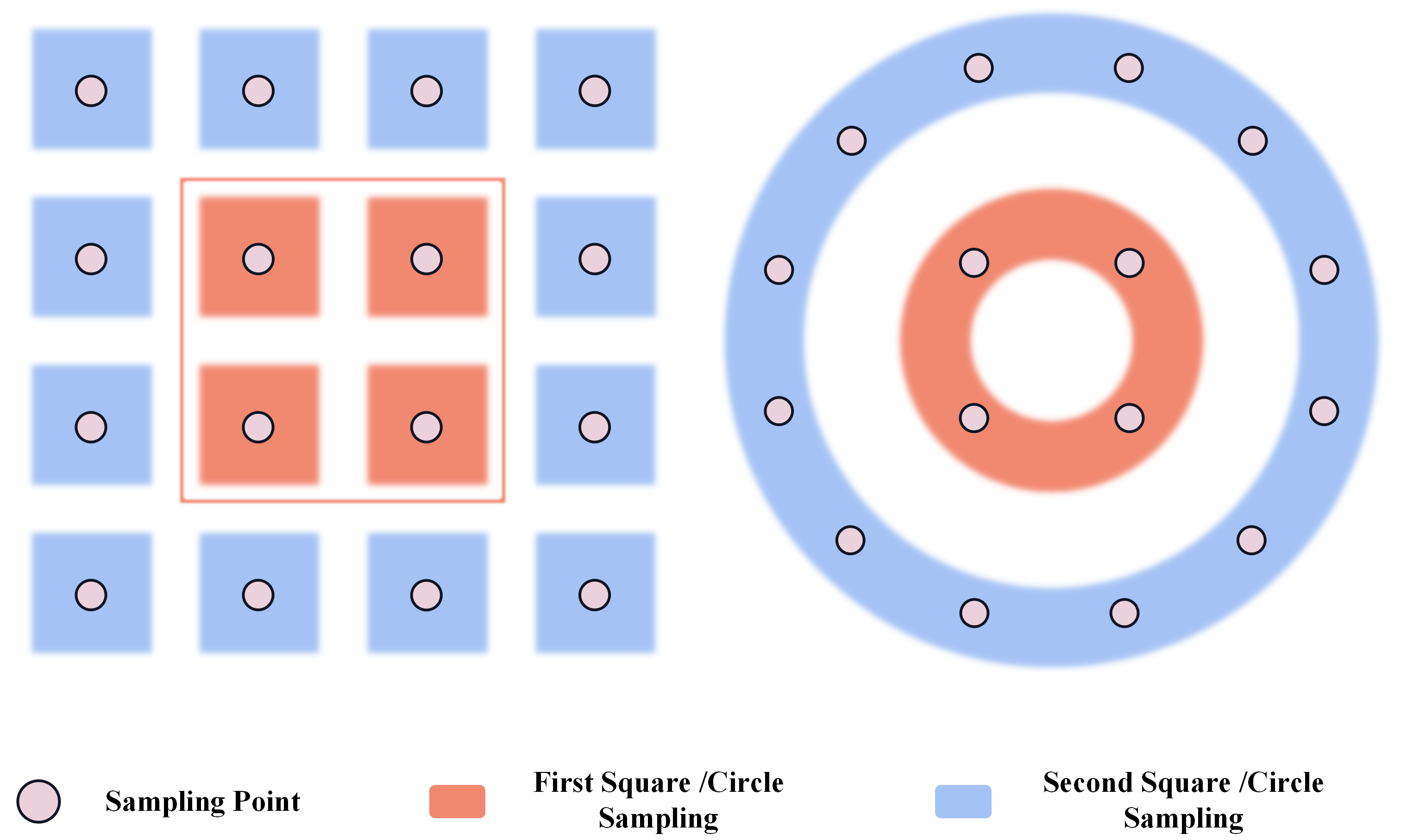}
    \caption{\textbf{Square Sampling and Circular Sampling}}
    \label{sample}
\end{figure}

Images captured with fisheye lens have a different pixel distribution from ordinary lens. Previous studies\cite{index-2,index-9} have indicated that this variation presents a particular structural pattern.

\begin{table}[htbp]
  \centering
  \caption{\textbf{Sampling Numbers}. $n$ represents the order of sampling from the center to the outer edge. This only shows the first four sample results.}
  \label{sample table}
  \begin{tabular}{ccc}
    \toprule
    n & Square Sampling Numbers & Circular Sampling Numbers \\
    \midrule
    1 & 4 & 4 \\
    2 & 12 & 12 \\
    3 & 20 & 20 \\
    4 & 28 & 28 \\
    \bottomrule
  \end{tabular} 
\end{table}

Inspired by Distortion Distribution Map (DDM)\cite{index-27}, we propose a circular sampling method. The typical sampling method in Transformer-based models is square sampling, Fig.\ref{sample} illustrates the differences between square sampling and circular sampling. In order to maintain the same overall sampling number, the number of circular sampling in each circle is the same as that of square sampling. As shown in Table\ref{sample table}, both of them take the same number of sampling points each time.

\subsection{Sector patch}

After circular sampling, the original patch method in the Transformer is no longer suitable. Therefore, we design a sector-shaped patch method. Each sector patch occupies the same area on the image, and the sampling points are uniformly sampled in each circle, which ensures that the number of sampling points in each sector patch is the same. Due to the quantity equivalence of circular sampling under SPE and square sampling under the original patch method, this ensures that the total number of sampling remains unchanged. However, compared to the original patch method, it can naturally filter out invalid pixels around the fisheye image. We represent the size of each sector patch in the following way. 
$$
(r_1, r_2, \theta, \alpha) \eqno{(5)}
$$
Where $r_1$ represents the radius of the inner circle of the sector, $r_2$ represents the radius of the outer circle, $\theta$ denotes the degree magnitude of the sector, and $\alpha$ indicates the angle of the sector's center with the r-axis in polar coordinates. Patches with the same radius from the center have the same sector shape. Combining the previous research\cite{index-9}, we figure that this patch method can adapt to the distortion pattern of the fisheye image.

\begin{figure*}[h]
    \centering
    \includegraphics[width=\textwidth]{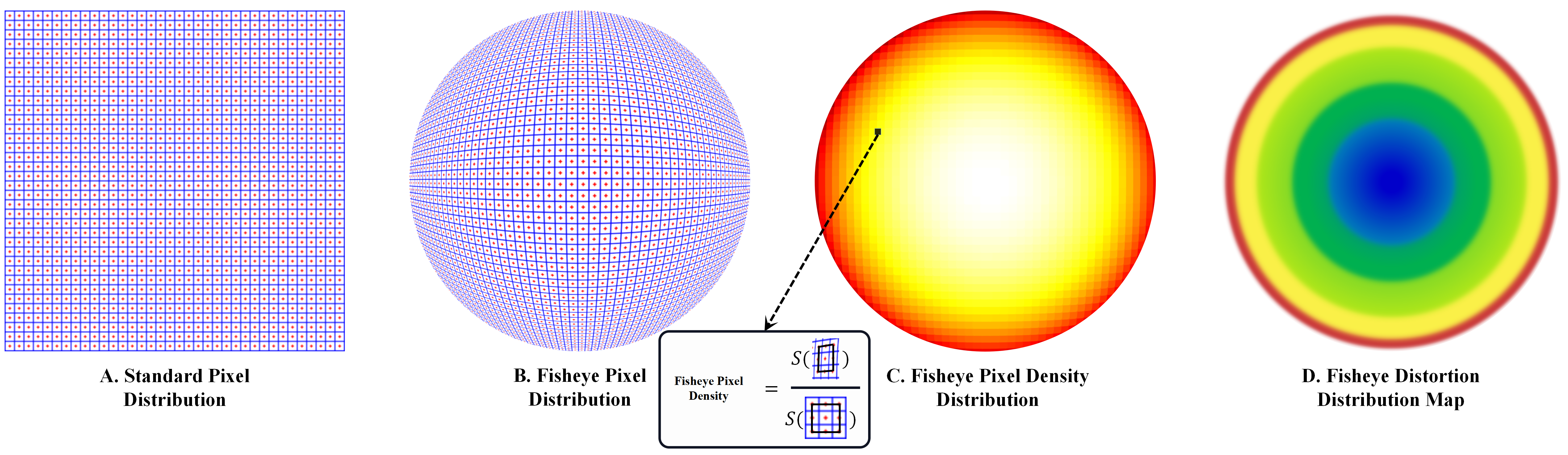}
    \caption{Image A represents the pixel distribution of the standard image. Image B is transferred from image A by fisheye transformation. Image C shows the pixel density distribution obtained by the pixel density calculation formula in Part IV. Image D is the DDM \cite{index-26}distortion distribution map of the fisheye image. $S(\cdot)$ represents the area.}
    \label{distributions}
\end{figure*}

\subsection{Polar position embedding}

In Transformer-based models, position embedding is added to the input sequence to encode the relative position of each token. Common approaches to position embedding are absolute position encoding, relative position encoding, and learnable position encoding. Since the sampled points on the fisheye image are represented by polar coordinates, we have designed a polar coordinate position encoding method.

$$
freq_{i}  =2^{(i-1) \times 2} / d_{model} \quad, \; i \in\left[1, d_{model}/ 2\right] \eqno{(6)}
$$

$$
P E_{i} = \left\{ \begin{array}{l}
\sin \left(r \times freq_{i}+\theta \times i\right)^{power},  \; i \; is \; even  \\
\cos \left(r \times freq_{i}+\theta \times i\right)^{power },  \; i \; is \; odd
\end{array} \right. \eqno{(7)}
$$

where tuple$(r, \theta)$ represents the position of each sampling point, $d_{model}$ denotes the dimension of the position encoding. The sine and cosine functions in the $i$-th dimension of the position encoding are modulated by a frequency parameter called $freq_i$. $PE_i$ is the formula for computing the position encoding, and the power parameter is used to adjust the output of the sine and cosine functions to enhance the expressive power of the position encoding.

\section{EXPERIMENTS AND RESULTS}

\subsection{Synthetic fisheye dataset}

We refer to and optimize the transfer method proposed by FDDB-360\cite{index-28}, and choose ImageNet-1K\cite{index-29} for producing synthetic fisheye images. Some examples are shown in Fig.\ref{detection}. we first shift the coordinate system from the top-left corner of the image to its center. This allows us to normalize the coordinates to a square with a double unit length. This new coordinate system is centered at $(0, 0)$, and the coordinates of the four corners are $(\pm1, \pm1)$ respectively. Subsequently, we map the square to a unit circle with:

$$
(x',y') = (x\sqrt{1-\frac{y^2}{2}},y\sqrt{1-\frac{x^2}{2}}) \eqno{(8)}
$$

Where $(x,y)$ are the coordinates of a point on the square and $(x',y')$ are the coordinates of the corresponding point on the unit circle.

Furthermore, in order to simulate the distortion of fisheye images at different focal lengths, we scale the coordinates on the unit circle:

$$
(x'',y'') = (x'e^{-kr^n},y'e^{-kr^n}) \eqno{(9)}
$$

Where $r =\sqrt{(x')^2+(y)'^2} $is the distance from point $(x',y')$  to the center. $k$ is a scaling factor and $n$ is an exponent factor.

We calculated the pixel distribution pattern under standard images and the pixel distribution pattern under fisheye images. Observation reveals that the structure of the image after the fisheye transformation is changed. As shown in Fig.\ref{distributions} B., the pixel density increases outward starting from the center. We figure that the pattern could be verified by calculating the pixel density distribution of the images after the fisheye transformation. In order to calculate the pixel density, we develop a method that uses a 3$\times$3 pixel square as the unit area in the standard image. For each pixel in the fisheye image, we calculate the area of a quadrilateral formed by the pixel point and its eight closest points and compare it with the corresponding area in the standard image. Due to some pixel loss after the fisheye transformation, we only selected pixels within a certain radius to calculate the density map. As shown in Fig.\ref{distributions} C. and D., our synthetic fisheye images demonstrate a pixel distribution that is in agreement with the DDM. 

\subsection{CUDA accelerated}

\begin{figure}
   \centering
   \includegraphics[width=0.48\textwidth]{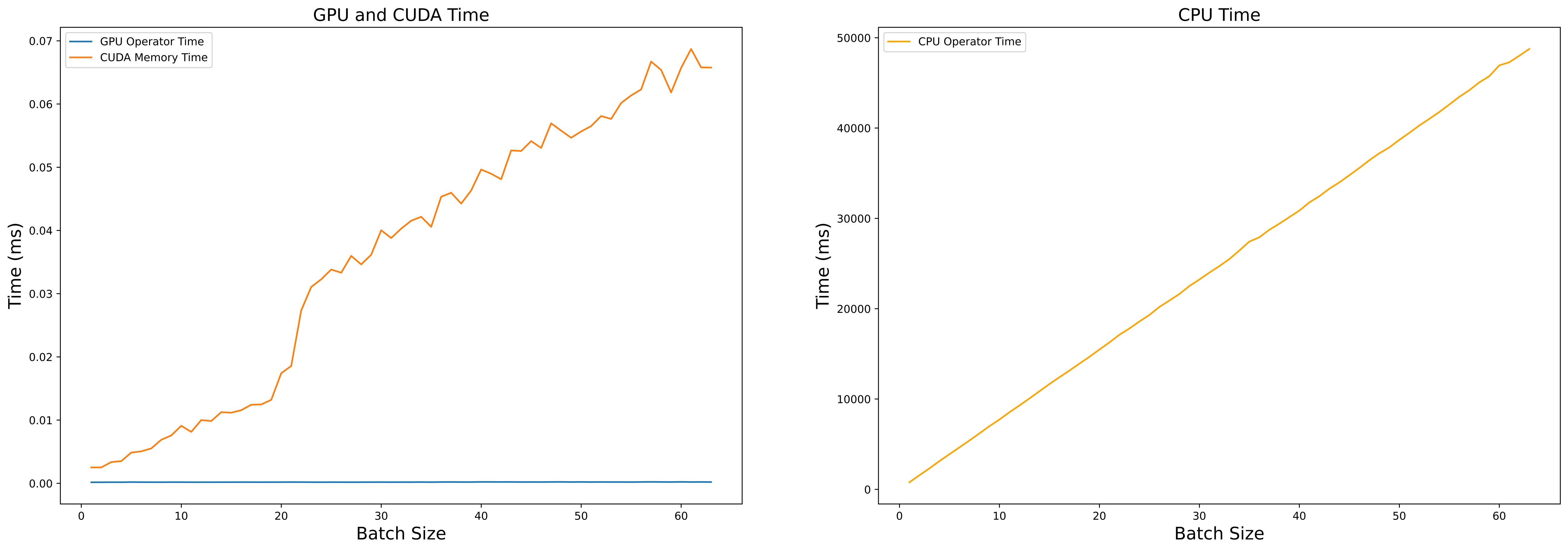}
   \caption{\textbf{Comparison of the time taken by CPU and GPU to generate Sector patches.} The left figure displays how the GPU operator's computation time and data import time vary with increasing batch size, using data with dimensions of $(B, 320, 320, 3)$. The right figure demonstrates the corresponding calculation time on the CPU.}
   \label{cuda}
\end{figure}

To the best of our knowledge, there's no algorithm that can efficiently generate our sector patches. Therefore, we need to create a new sampling algorithm. Considering that the entire algorithm is a pixel-wise operation both in sampling and patch generation, which means traversing all sampling points. Therefore, we use CUDA for acceleration. It significantly improves the algorithm's efficiency. Specifically, we use CUDA for parallel allocation and computation of sampling points. The high memory access capability of GPUs helps us improve the algorithm's data access speed. In this way, we can generate a large number of patches in a short period and ensure that the sampling points are correct. Additionally, we compare it with the algorithm for generating sector patches on the CPU, and the results are shown in Fig.\ref{cuda}. Our operator significantly improves the efficiency of sampling and generating patches, and shows a stable performance at the millisecond level.

\subsection{Semantic feature extraction of fisheye image}

\begin{table}[htbp]
 \centering
 \caption{\textbf{Image classification performance on the Fisheye ImageNet-1K validation set.}  SPVT represents the PVT model with the SPE module and SViT represents the ViT model with the SPE module. PPE is the abbreviation for Polar Patch Embedding. We present further details about our experiment settings in the $Experiment$ $settings$ section.}
 \label{result}

 \begin{tabular}{ccc}

   \toprule
   Method & \#Param(M) & Top-1 Acc(\%) \\
   \midrule
   PVT\_Small & 24.5 & 51.76 \\
   SPVT\_Small & 24.5 & 54.56 \\
   PVT\_Medium & 44.2 & 52.85\\
   SPVT\_Medium & 44.2 & 55.76 \\
   SPVT\_Small+PPE & 24.4 & 55.09 \\
   ViT\_Base\_16 & 52.61 & 60.84 \\
   SViT\_Base\_16 & 52.61 & 61.59 \\ 
   \bottomrule
 \end{tabular} 
\end{table}

We conducted two experiments to test whether SPE modules can enhance the semantic feature extraction capacity of ViT and PVT on the Fisheye ImageNet-1k dataset. The models were trained and evaluated using top-1 accuracy as the metric. The results are shown in Table\ref{result}. ViT achieves 60.84@top-1, while PVT\_Mediun achieves 52.85@top-1 on the validation set. We think the gap between the two is mainly because PVT's Spatial Reduction Attention reduces the calculation amount compared to ViT's Multi-Head Attention. This diminishes the model's ability to perceive global and local distortions. After embedding ViT and PVT with the SPE module, the classification top-1 accuracy improved by 0.75 and 2.8, respectively. If we add polar position embedding to SPVT, the accuracy can increase by 0.52. These results demonstrate that the SPE module can enhance the semantic feature extraction capacity of fisheye images.

\begin{figure}[h]
\centering
\includegraphics[width=0.48\textwidth]{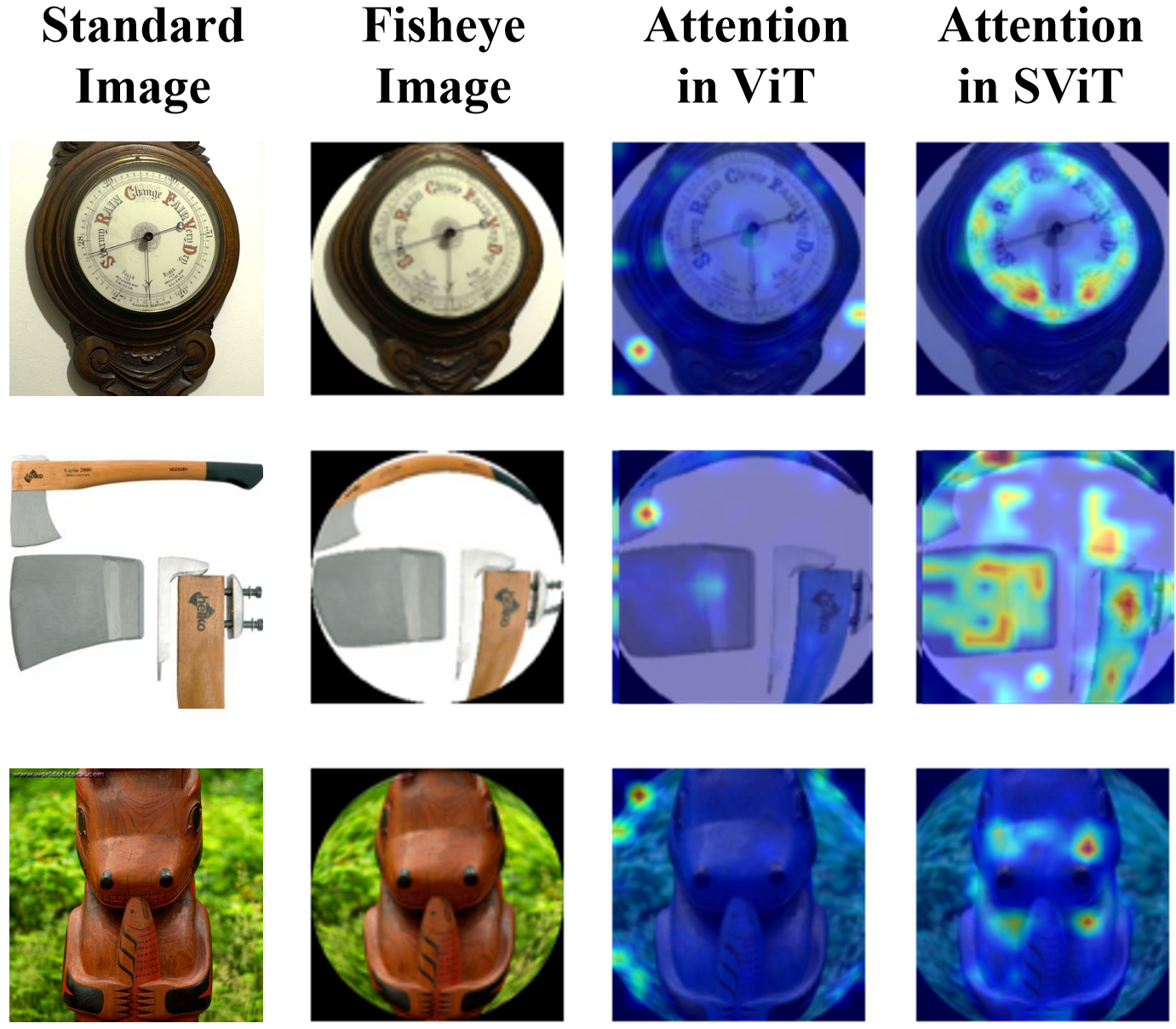}
\caption{\textbf{Example of synthetic fisheye images and comparison of attention in ViT and SViT}}
\label{attention}  
\end{figure}

\begin{figure*}
   \centering
   \includegraphics[width= \textwidth]{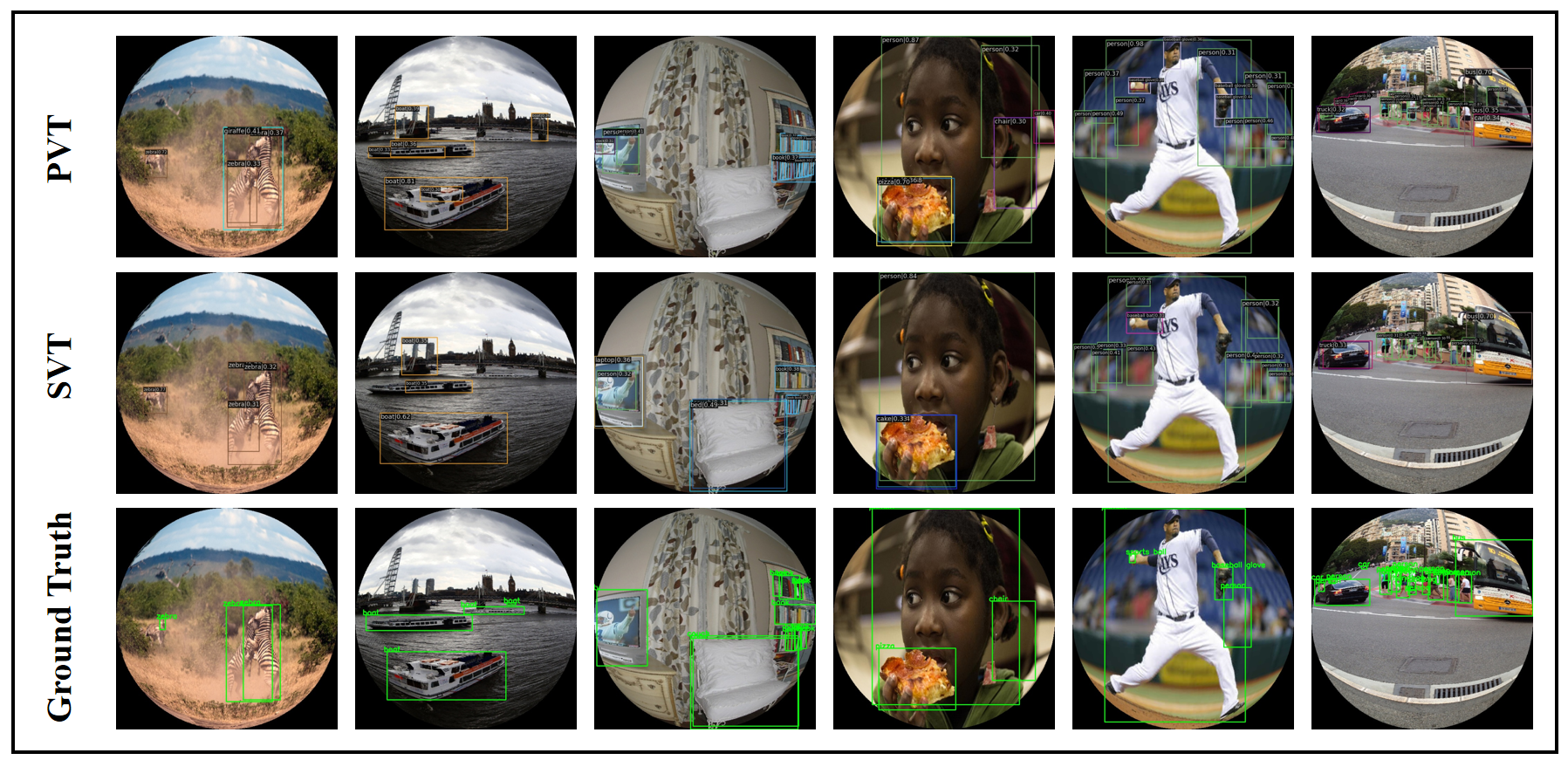}
   \caption{ \textbf{Ground truth and object detection results of PVT and SPVT.}}
   \label{detection}
\end{figure*}

We show the attention distribution of ViT and SViT on images in Fig.\ref{attention} with the Transformer model's attention visualization tool provided by Chefer\cite{index-30}. SViT has better visual semantic extraction capabilities, indicating that its design is better for learning the distortion pattern of fisheye images. SPE helps the Transformer model capture the key features of fisheye images, allowing it to focus more accurately on essential areas using attention distribution.

In addition, we explore the detection performance of PVT and SPVT on COCO-fisheye, and the result is shown in Fig.\ref{detection}

\subsection{Experiment Settings}
The experimental configuration is designed based on the Fisheye Imagenet-1K dataset, consisting of 1.28 million training images and 50,000 testing images from 1,000 categories. The optimization algorithm used is AdamW with a learning rate of 0.0001, betas of (0.9, 0.999), and weight decay of 0.05. The batch size is set to 64. And the experiments are deployed on four RTX3090 GPUs for 100 epochs. The learning rate is adjusted using a cosine schedule\cite{index-31}.

The first model used is PVT\_small, which is the original model configuration described in\cite{index-25}. For the second model, we used a variant of the ViT architecture. This ViT variant, which we named ViT\_Base\_16, has a backbone network with patch size of 16, a depth of 6, 16 attention heads, a dimension of 1024, a MLP dimension of 2048, a dropout rate of 0.1, and an embedding dropout rate of 0.1.

\section{CONCLUSIONS}

In this paper, we propose a sector patch method with an effective sampling way, denoted as Sector Patch Embedding(SPE). Based on the expriments presented, the following conclusion can be drawn:

(1) SPE is proved to be effective. We conduct experiments to validate it on both uniform-scale and multi-scale Transformers. The results demonstrate that SPE can improve the feature extraction capability of Transformers. SPE can be integrated into most Transformer-based models.

(2) Polar Position Embedding(PPE) is also effective in fisheye feature extraction. The model's performance is improved after adopting PPE.

Regarding future work, we plan to optimize the performance of SPE on fisheye object detection. Furthermore, finetuning the SPE to adapt to different degrees of distortion is also a worthy task.

\section*{ACKNOWLEDGMENT}
This work was partly supported by National Natural Science Foundation of China (Grant No. U1913203, 61973034, 62233002 and CJSP Q2018229).
The authors would like to thank Xi Xu, Tianji Jiang, Zhaoxiang Liang, Xihan Wang, and all other members of ININ Lab of Beijing Institute of Technology for their contribution to this work.


\bibliographystyle{ieeetr}
\bibliography{reference}

\end{document}